\PassOptionsToPackage{normalem}{ulem}
\PassOptionsToPackage{hyperfootnotes=false}{hyperref}
\documentclass{TEAI}
\usepackage{helvet}
\usepackage[nomath]{lmodern}
\usepackage{makecell}
\usepackage{tabularx}
\usepackage{enumitem}
\hypersetup{
  pdftitle={Prism: Hierarchical Text Detection with Minimalist Design},
  pdfauthor={Yuchen Su, Xuesheng Wang, Zhilong Ji, Jinfeng Bai, Shijian Lu, Zhineng Chen},
  pdfkeywords={Hierarchical Text Detection, Minimalist Design, Geometric Inclusion}
}

\makeatletter
\newcommand{\useembeddedpdfgroup}{%
  \ifdefined\Gread@transgroupfalse
    \let\Gread@transgrouptrue\Gread@transgroupfalse
  \fi}
\makeatother

\newcolumntype{Y}{>{\centering\arraybackslash}X}
\colorlet{resultrow}{seedblue!7}
\newcommand{\resulttableformat}{%
  \small
  \setlength{\tabcolsep}{2.5pt}%
  \renewcommand{\arraystretch}{1.14}%
  \setlength{\heavyrulewidth}{0.7pt}%
  \setlength{\lightrulewidth}{0.4pt}%
  \setlength{\cmidrulewidth}{0.32pt}%
  \setlength{\aboverulesep}{1.8pt}%
  \setlength{\belowrulesep}{2pt}%
}

\definecolor{c_red}{HTML}{ea4335}
\definecolor{c_green}{HTML}{34a853}
\colorlet{tablered}{c_red!76!black}
\colorlet{tablegreen}{c_green!70!black}
\newcommand{\resultdelta}[3]{%
  \,\textcolor{#1}{\raisebox{.37ex}{\fontsize{7}{8}\selectfont\ensuremath{#2}#3}}}
\newcommand{\qincnew}[1]{\resultdelta{tablegreen}{\uparrow}{#1}}
\newcommand{\qdecnew}[1]{\resultdelta{tablered}{\downarrow}{#1}}
\newcommand{\qeqnew}[1]{\resultdelta{black!63}{\uparrow}{#1}}

\title{Prism: Hierarchical Text Detection with \\ Minimalist Design}

\author{%
  \spaceskip=.21em plus .04em minus .04em\relax
  Yuchen Su\textsuperscript{1,2}, Xuesheng Wang\textsuperscript{1,2}, Zhilong Ji\textsuperscript{3},
  Jinfeng Bai\textsuperscript{3}, Shijian Lu\textsuperscript{4}, Zhineng~Chen\textsuperscript{1,2,\textdagger}%
}
\affiliation{%
  \textsuperscript{1}Institute of Trustworthy Embodied AI, Fudan University,\\
  \textsuperscript{2}Shanghai Key Laboratory of Multimodal Embodied AI,\\
  \textsuperscript{3}Tomorrow Advancing Life,
  \textsuperscript{4}Nanyang Technological University\\

}

\abstract{
Hierarchical text detection aims to localize words, text lines, and paragraphs while parsing their structural relationships. It provides essential structural cues for visual text understanding. To maximize accuracy, existing methods typically rely on complex multi-scale feature fusion necks and heavy-weight heads, resulting in bloated architectures and low inference efficiency. In this paper, we argue that a sufficiently pre-trained backbone can effectively replace these sophisticated external modules, while simple geometric inclusion suffices for hierarchical text structural parsing. Thus, we propose Prism, a minimalist framework for hierarchical text detection. Prism performs multi-granularity detection by attaching parallel lightweight heads to a single-scale backbone feature map, covering words, text lines, and paragraphs, and reconstructs the structural hierarchy via an efficient geometric inclusion algorithm. In addition, we systematically compare representative lightweight head archetypes and show that pixel-level regression offers the best accuracy-efficiency trade-off in this setting. 
Experiments on multiple public benchmarks show that our significantly simplified architecture achieves state-of-the-art (SOTA) performance with remarkable efficiency. Specifically, it runs $1.7\times$ faster than LRANet on CTW1500 while achieving higher accuracy, and surpasses the current SOTA method, OmniParser V2, by over 6 PQ points on the hierarchical HierText dataset.
Code will be released soon.
}
\correspondence{\email{zhinchen@fudan.edu.cn}}

\begin{document}
% The class title block is followed by the Introduction's own heading space.
\mymaketitle
\begingroup
  \renewcommand{\thefootnote}{\textdagger}
  \makeatletter
  \renewcommand{\@makefntext}[1]{\noindent\@makefnmark\,#1}
  \makeatother
  \footnotetext{Corresponding author.}
  \enlargethispage{\dimexpr\skip\footins+\footnotesep\relax}
\endgroup
\suppressfloats[t]

\section{Introduction}
% \IEEEPARstart{H}{ierarchical} text detection aims to localize visual text across different granularities (\textit{e.g.}, words, lines, and paragraphs) and parse their topological inclusion relationships. This structured parsing is crucial for a comprehensive understanding of texts in both natural scenes and electronic documents. However, an efficient and accurate hierarchical text detection framework remains lacking, limiting its applicability in many real-world scenarios.

\begin{figure}[t]
    \centering
    \begin{minipage}{.96\linewidth}
        \centering
        \useembeddedpdfgroup
        \includegraphics[width=\linewidth]{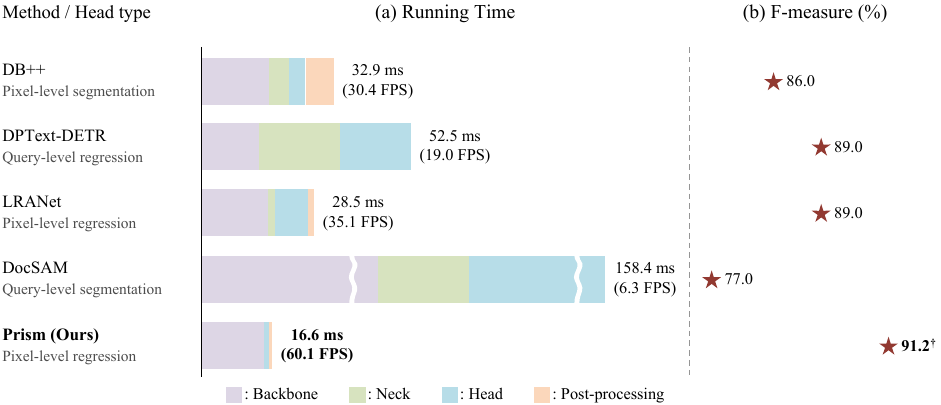}
        \caption[Comparison between Prism and other mainstream text detection methods on Total-Text.]{\textbf{Comparison between Prism and other mainstream text detection methods on Total-Text.} (a)~shows
        a~breakdown of the inference time across different components for each method at a uniform input resolution of 1024 $\times$ 1024. (b)~compares their F-measures, evaluated at each method's optimal resolution. ${\dagger}$~denotes our method adopts sufficient pre-training.}
        \label{fig:efficient-accurate}
    \end{minipage}
\end{figure}

Text detection aims to localize textual instances in images. Hierarchical text detection goes one step further by detecting text at multiple granularities (\textit{e.g.}, words, lines, and paragraphs) and parsing their structural relationships. Such structured outputs provide richer layout information for text-rich natural scenes and electronic documents \cite{du2025svtrv2, du2025unirec}. However, an efficient and accurate hierarchical text detection framework remains lacking, limiting its applicability in many real-world scenarios.

To pursue higher accuracy, existing text detection methods typically adopt an ``architectural stacking'' paradigm at the expense of efficiency. Specifically, mainstream solutions rely on increasingly elaborate designs: 1) \textbf{Complex Necks}--the feature pyramid network (FPN) \cite{lin2017feature} and its variants (\textit{e.g.}, PANet \cite{liu2018path}, BiFPN~\cite{tan2020efficientdet}) have become the standard for handling multi-scale objects, and complex multi-scale deformable attention modules \cite{zhu2020deformable, ye2023dptext} have also been widely adopted; 2) \textbf{Heavy-weight Heads}--bottom-up methods leverage computationally intensive high-resolution heads for precise pixel-level prediction \cite{liao2022real,zhao2024cbnet,ye2024hi}, while top-down methods use multiple parallel heads \cite{zhu2021fourier,liu2021abcnet,su2026lranet++} or global attention modules \cite{bu2024srformer,su2025explicit,wan2024omniparser} to handle multi-scale text instances. 
Furthermore, for hierarchical text detection, complex relation-reasoning modules, such as the explicit word-to-line and line-to-paragraph relationship modeling in LayoutFormer \cite{liang2024layoutformer},
are typically needed to parse the structural inclusion across different text granularities. 

% We depart from this complexity-first paradigm and instead adopt a minimalist design philosophy, aiming for effective hierarchical text detection with a simple network architecture. Our design is based on three observations: 

We depart from this complexity-first paradigm and instead revisit text detection from a minimalist perspective, exploring whether a simple architecture can be sufficient when backed by strong pre-training. In general object detection, 
plain ViT \cite{dosovitskiy2020image} backbones pre-trained with MAE \cite{he2022masked} have been shown to achieve competitive performance with simple multi-scale adaptation, as demonstrated by ViTDet \cite{li2022exploring}. This raises a natural question: can text detection be simplified even further? We base our minimalist design on three task-specific observations: 1) Text detection fundamentally targets a single foreground category composed of a finite set of structural primitives (\textit{e.g.}, strokes and characters). We thus hypothesize that a single-scale feature map from a sufficiently powerful, extensively pre-trained backbone is discriminative enough to render multi-scale fusion necks like FPN redundant. 2) Text detection is inherently a purely perceptual task that requires only intrinsic visual features of each text instance, without the need for complex semantic reasoning (\textit{e.g.}, differentiating headings from body text). Therefore, lightweight heads are likely sufficient for the final prediction. 3) The spatial layout of text inherently encodes its hierarchical inclusion. Thus, simple geometric calculations can naturally replace complex word-line-paragraph relation modules.

\begin{figure}[t]
    \centering
    \begin{minipage}{.92\linewidth}
        \centering
        \includegraphics[width=\linewidth]{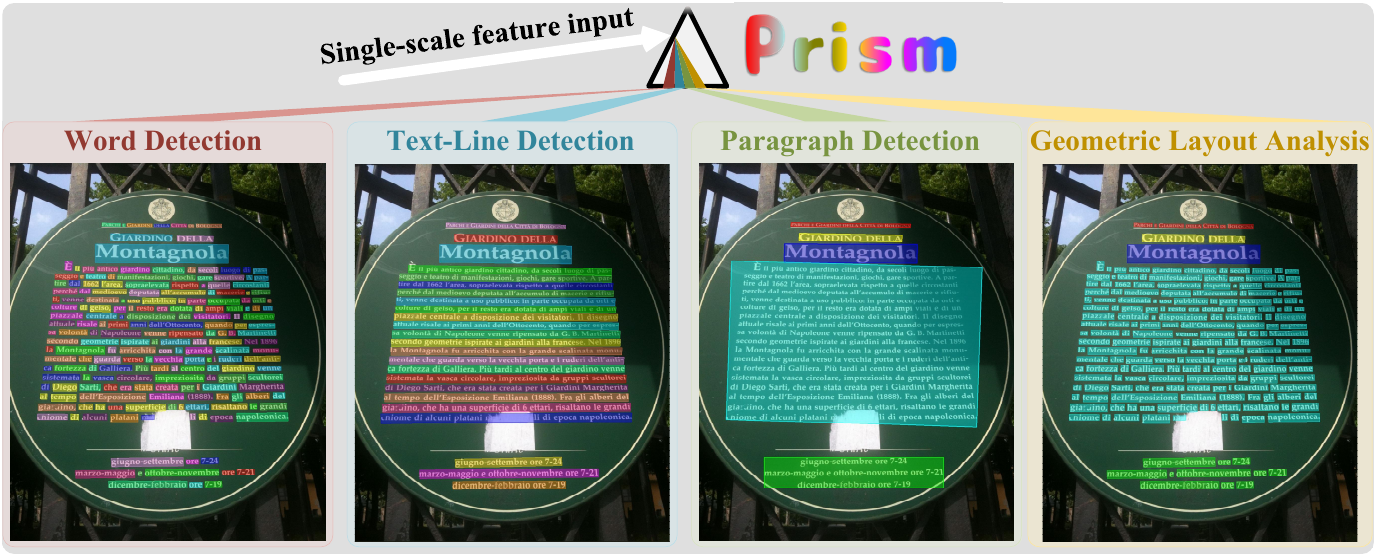}
        \caption{\textbf{Visualization of Prism.} Like a prism refracting light, our framework decomposes a single-scale feature into a rich spectrum of hierarchical outputs, spanning words, lines, paragraphs, and their geometric layouts.}
        \label{fig:wlpg}
    \end{minipage}
\end{figure}

Building upon these insights, we propose Prism, a minimalist hierarchical text detection framework. True to its name, just as a prism refracts a single beam of light into a vibrant spectrum, Prism decomposes the single-scale feature into a diverse set of structural outputs, as illustrated in Fig.~\ref{fig:wlpg}. Specifically, it consists solely of a backbone and a set of parallel, lightweight expert heads built upon a single-scale feature map. Conceptually, this resembles a multi-expert architecture, with each head specializing in a specific task (\textit{e.g.}, word or line detection). Finally, we employ a box-approximation geometric inclusion algorithm to efficiently and accurately parse the hierarchical structure. 

\enlargethispage{.55\baselineskip}
To fully realize this minimalist framework, the design of the lightweight head is crucial. Thus, we categorize and systematically analyze lightweight implementations of four head archetypes: pixel-level regression (\textit{e.g.},~YOLO-style~\cite{wang2024yolov10}), query-based regression (\textit{e.g.}, DETR-style \cite{carion2020end}), pixel-level segmentation (\textit{e.g.}, DB-style \cite{liao2020Real}), and query-level segmentation (\textit{e.g.}, MaskFormer-style \cite{cheng2021per}). Our study shows that a pixel-level regression head, which explicitly leverages spatial priors and perceives instances holistically, achieves the best balance of accuracy and efficiency.

Owing to its minimalist design, Prism yields remarkable inference speed. As shown in Fig.~\ref{fig:efficient-accurate}, Prism is $1.7\times$ faster than the leading real-time detector LRANet \cite{su2024lranet} while maintaining superior accuracy. Moreover, on the hierarchical dataset HierText \cite{long2022towards}, our method surpasses the SOTA Omniparser V2 \cite{yu2026omniparser} by at least 6\% in Panoptic Quality (PQ). 

The main contributions of this work are as follows:

\begin{itemize}[leftmargin=1.45em,labelsep=.52em,topsep=3pt,itemsep=3pt,parsep=0pt]
\item We introduce a minimalist design paradigm for unified hierarchical text detection, demonstrating that as backbone models and pre-training data scale up, complex multi-scale necks and heavy heads become increasingly redundant.

\item We propose Prism, a simple framework that couples single-scale multi-granularity detection with parameter-free geometric inclusion, enabling efficient word-line-paragraph hierarchy parsing without multi-scale necks or learnable relation modules.
% $\bullet$ We propose Prism, a simple framework that employs parallel, granularity-specific lightweight heads to a single-scale feature map for multi-granularity text detection, and utilizes a parameter-free geometric inclusion algorithm to efficiently parse the final hierarchical layout.

\item Extensive experiments over multiple challenging benchmarks show that Prism achieves superior detection accuracy and computational efficiency despite its significantly simplified architecture.
\end{itemize}

\section{Related Work}

Text detection has evolved from conventional document-oriented scenarios to more challenging in-the-wild settings, where text instances often exhibit arbitrary shapes, large scale variations, cluttered backgrounds, and complex illumination. According to the representation of text instances, existing methods can be broadly grouped into segmentation-based and regression-based approaches. We review these two families first, then discuss hierarchical text detection and recent minimalist detection designs that motivate our work.

\subsection{Segmentation-Based Text Detection}

Segmentation-based methods, mainly inspired by instance segmentation \cite{bolya2019yolact,wang2020solov2,chen2020blendmask,cheng2021per}, treat text detection as a segmentation problem. These methods have evolved along two primary paradigms: bottom-up pixel-level classification and top-down query-level mask prediction.

\textbf{Pixel-level segmentation} approaches typically predict dense pixel-wise segmentation maps, which are then grouped into individual text instances. For example, 
PSENet \cite{wang2019shape} reconstructs instance contours using a multi-scale kernel expansion strategy, while TextPMs \cite{zhang2022arbitrary} uses an iterative region growing algorithm. DB series \cite{liao2020Real,liao2022real} introduce a groundbreaking differentiable binarization module to simplify post-processing.
KAC \cite{zheng2024kernel} proposes a kernel adaptive convolution module to enhance segmentation results.

However, computationally intensive post-processing remains a bottleneck. For example, as shown in Fig.~\ref{fig:efficient-accurate}, the post-processing stage of the highly-optimized DB++ \cite{liao2022real} still accounts for more than 20\% of the total inference time.

\textbf{Query-level segmentation} approaches address the aforementioned limitations from a top-down perspective. Following the instance segmentation method MaskFormer \cite{cheng2021per}, they use a set of learnable object queries to predict the overall mask for each instance, and thus avoid the complex post-processing. For instance, DocSAM \cite{li2025docsam} introduces semantic queries to interact with standard instance queries to handle heterogeneous text detection, segmentation, and layout analysis tasks. 

However, while simplifying the post-processing, the quality of mask prediction depends heavily on complex multi-layered Transformer decoders and multi-scale fusion necks. 

\subsection{Regression-Based Text Detection}

Regression-based methods represent text instances as parameterized vectors for regression. Based on the positive sample assignment strategy, they can be divided into two main paradigms: pixel-level regression and query-level regression.

\textbf{Pixel-level regression} approaches typically perform pixel-by-pixel regression on feature maps and then use non-maximum suppression (NMS) to filter redundant prediction results. 
For example, ABCNet \cite{liu2020abcnet} adopts the Bernstein polynomial to transform the long sides of the text into Bezier curves, and performs regression using the FCOS \cite{tian2019fcos} architecture. To fully exploit text shape relevance and reduce post-processing time, LRANet \cite{su2024lranet} proposes low-rank approximate text shape representations and a dual allocation strategy for positive samples.

\textbf{Query-level regression} approaches, pioneered by DETR \cite{carion2020end}, frame parameterized text shape regression as a set prediction problem with sparse positive samples, eliminating the need for NMS. For instance, DPText-DETR \cite{ye2023dptext} proposes an explicit point-coordinate query modeling method to handle the varying scales and shapes of text instances. SRFormer \cite{bu2024srformer} further introduces a segmentation branch in shallow layers to provide pixel-level priors.

Although they eliminate NMS, this does not inherently lead to superior efficiency, as the complexity is transferred to the heavy-weight Transformer decoder itself, which must perform a similar function of redundancy removal through its self-attention modules.

\subsection{Hierarchical Text Detection}

Hierarchical text detection aims to detect multi-granularity text and then perform geometric layout analysis. UniDec \cite{long2022towards} first proposes unifying scene text detection and geometric layout analysis, using a query-level segmentation-based method to predict words, lines, and paragraphs, and an affinity matrix to determine paragraph grouping. LayoutFormer \cite{liang2024layoutformer} introduces a unified framework with three parallel Transformer decoders (for words, lines, and paragraphs, respectively), complemented by modules that explicitly model word-to-line and line-to-paragraph relationships. DAT \cite{wan2024towards} proposes a cross-granularity interactive attention module for multi-granularity text detection. Moreover,
OmniParser V2 \cite{yu2026omniparser} frames the task as a sequence generation problem, generating complete hierarchical relationships by inserting line and paragraph tags into its structured point sequences.

\begin{figure}[t]
    \centering
    \begin{minipage}{\linewidth}
        \centering
        \includegraphics[width=\linewidth]{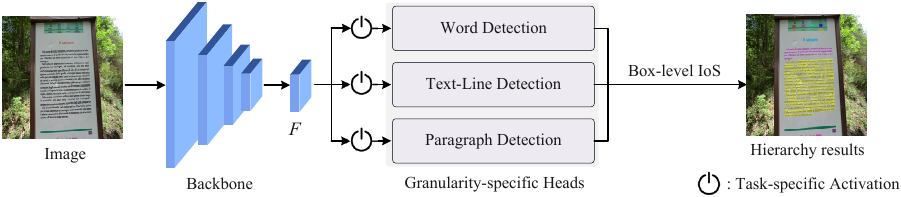}
        \caption{\textbf{The overview of our Prism}. It consists of only two main components: a backbone and a set of parallel, lightweight heads. ``Task-specific activation'' denotes selectively activating a specific expert head based on the required task configuration.}
        \label{fig:overview}
    \end{minipage}
\end{figure}

However, these approaches typically rely on complex multi-scale necks and heavy-weight heads, as well as intricate hierarchical modeling modules to reason about the relationships across different text granularities, leading to low efficiency, which is a core issue our work seeks to challenge.

\FloatBarrier
\section{Methodology}

\subsection{Minimalist Design Philosophy}

Current text detection methods typically follow  general object detection or segmentation paradigm, relying on multi-scale fusion necks to adapt to varying text scales and shapes, coupled with increasingly heavy-weight heads to refine predictions. However, this ``complexity-first'' consensus overlooks the first principles of text detection. Unlike general object detection, it is a single-category task without inter-class confusion, and image text is a structured visual pattern composed of a finite set of primitives such as strokes and characters.

Based on these first principles, we hypothesize that these elaborately designed necks and heavy heads are essentially crutches devised to mitigate the weak representational power of less capable backbones or those with limited pre-training. With the emergence of more powerful backbones and the availability of large-scale pre-training, this complexity may no longer be necessary. Thus, a text detection model requires only two indispensable components: an extensively pre-trained backbone for extracting generic features and a set of heads for mapping those features to specific task outputs. Furthermore, for hierarchical text detection, learnable relation reasoning modules are unnecessary, as natural geometric calculations of spatial relationships are sufficient to parse the structural layout.

This hypothesis presents a critical design choice: whether to integrate these task heads into the backbone or to separate them. We argue strongly for their separation. A task-agnostic backbone can learn more universal representations, while separate, task-specific heads provide crucial modularity and flexibility. This decoupled design supports features like on-demand task activation and avoids the need to retrain the entire model when adding new tasks. Thus, our architectural blueprint is a single backbone followed by parallel, lightweight, and independent heads for different text granularities.

\subsection{Overall Framework}

\begin{figure}[t]
    \centering
    \begin{minipage}{\linewidth}
        \centering
        \useembeddedpdfgroup
        \includegraphics[width=\linewidth]{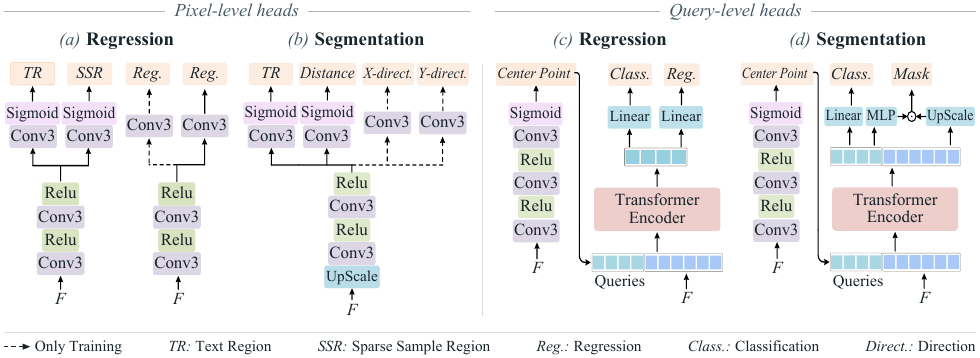}
        \caption{\textbf{Detailed architectures of the four lightweight head archetypes.}
        %(a) is adopted as the default parallel Expert Head for Prism.
        }
        \label{fig:head}
    \end{minipage}
\end{figure}

Following the design philosophy outlined above, the overall framework of Prism is radically simple, as illustrated in Fig.~\ref{fig:overview}. 
Prism utilizes only the single-scale feature map from a single deep stage of the backbone as the unified entry point for all subsequent tasks. True to its name, the framework operates like an optical prism: it maps this single feature representation to a spectrum of structured outputs through parallel lightweight heads. We refer to these granularity-specific heads as Expert Heads, as each one is responsible for a particular output level, such as word, text-line, or paragraph detection.
Furthermore, a task-specific activation mechanism selectively routes features to the required heads, enabling flexible on-demand inference and targeted training on datasets with different annotations.

Notably, the selection of an appropriate output feature scale is key to the success of this minimalist design. Standard hierarchical backbones produce feature maps at various downsampling rates (\textit{e.g.}, 4x, 8x, 16x, 32x). Deeper features (\textit{e.g.}, 32x), while semantically rich, have a low spatial resolution that is detrimental to localizing small text instances. Conversely, shallower features (\textit{e.g.}, 4x or 8x) significantly increase the computational overhead of the subsequent prediction heads due to their large spatial size. Therefore, we adopt the 16x downsampled feature map (\textit{e.g.}, the C4 stage of ResNet) as the sole feature source. This scale strikes an ideal balance between semantic abstraction and spatial resolution, while also allowing for the truncation of deeper backbone stages to further reduce computational cost without compromising performance.

Finally, we adopt simple geometric calculations to parse hierarchical relationships. However, iterative intersection operations on irregular text contours are highly time-consuming, especially for dense predictions. To ensure high efficiency, we approximate the text instances with their minimum bounding boxes and calculate their intersection-of-sets (IoS). Crucially, rather than relying on complex polygon intersections, we reconstruct the structural hierarchy through a bottom-up sequence of vectorized box-level IoS calculations, establishing the layout (from words to lines, then lines to paragraphs) with remarkable efficiency.

\subsection{Lightweight Head Archetypes}
\label{method:head}

% As mentioned above, the parallel Expert Heads (\textit{e.g.}, for word, line, and paragraph detection) share an identical architectural design to ensure simplicity. To determine the optimal architecture for these shared heads, we systematically built and explored four distinct archetypes corresponding to the mainstream detection paradigms, detailed below.

To identify a suitable lightweight head design for the proposed single-scale framework, we systematically explore four representative archetypes corresponding to mainstream detection paradigms.

\noindent \textbf{Pixel-level regression head.}
As shown in Fig.~\ref{fig:head}(a), this head consists of two parallel branches: a classification branch and a regression branch. Each adopts two $3\times3$ convolutional layers to extract task-specific features. Following the dual-path scheme in \cite{su2024lranet}, both branches are split into a dense text region (TR) path and a sparse sample region (SSR) path. The dense TR path provides ample supervision signals by treating all locations within the ground-truth text region as positive samples for contour coefficient regression, ensuring robust feature learning. In contrast, the sparse SSR path is dedicated to accelerating inference. Its positive samples are determined via the Hungarian algorithm, which strictly assigns one positive sample per instance to eliminate the time spent filtering redundant predictions.
For each granularity, the same regression branch and assignment strategy are applied independently, while the prediction targets are derived from the corresponding word-, line-, or paragraph-level annotations.

\noindent \textbf{Pixel-level segmentation head.} 
As shown in Fig.~\ref{fig:head}(b), this head first upscales the input features by 4$\times$ to recover spatial details, followed by two $3\times3$ convolutional layers for refinement. Following \cite{zheng2024kernel}, we predict four parallel maps: a text segmentation (TR) map, a distance map, and two pixel-wise direction maps (X and Y). The direction maps serve as auxiliary supervision during training to separate adjacent instances. During inference, only the TR and distance maps are utilized. Specifically, the text center region is extracted by thresholding the distance map, while the TR map acts as auxiliary information to filter false positives. Finally, these preserved centers are expanded to reconstruct the text contours.

\noindent \textbf{Query-level regression head.}
As shown in Fig.~\ref{fig:head}(c), this head adopts an encoder-only DETR-like paradigm. It first employs a lightweight convolutional branch to predict the top-$k$ (\textit{e.g.}, top-100) most salient center points from the feature map $F$. These points are then encoded as a set of learnable queries with discriminative positional cues. To simplify the structure, these queries are concatenated with the flattened feature map $F$ to form a single, composite input sequence, enabling a ``dual-path'' optimization within a unified Transformer Encoder consisting of three layers. Specifically, unlike conventional DETR-like structures that utilize an encoder for visual features followed by a decoder for query refinement, our design achieves both functions within the single encoder. This structure allows the queries to attend to the full context of $F$ while simultaneously refining $F$'s own representations, effectively achieving an implicit cross-attention. This strategic simplification eliminates the need for a separate, heavyweight Transformer Decoder. Finally, the output features corresponding to the refined queries are fed into two parallel linear layers to generate the final classification and regression results.

\noindent \textbf{Query-level segmentation head.}
As shown in Fig.~\ref{fig:head}(d), this head adopts an encoder-only paradigm similar to that of our query-level regression head (Fig.~\ref{fig:head}(c)). The key distinction lies in the output stage, where the regression branch is replaced by a MaskFormer-style segmentation module. Specifically, the refined query features are fed into a Linear layer for classification and an MLP to generate mask embeddings. Concurrently, the refined visual features are upscaled by 4$\times$ to serve as high-resolution pixel embeddings. The final segmentation masks are generated via the dot product of the query-specific mask embeddings and the pixel embeddings. Notably, to accelerate convergence, the convolutional center-point prediction branches in both query-level heads (Fig.~\ref{fig:head}(c)(d)) are initialized with pre-trained weights from the pixel-level segmentation head.

Accordingly, we use the pixel-level regression head in Prism, with detailed quantitative justification provided in Sec.~\ref{ablation}.

% We adopt the pixel-level regression head (Fig.~\ref{fig:head}(a)) as our default Expert Head. Detailed quantitative analysis justifying this choice is provided in Sec.~\ref{ablation}.

\subsection{Optimization and Post-processing}

% To effectively train our unified model on diverse datasets that often contain incomplete annotations (\textit{e.g.}, some datasets provide only word-level labels, while others provide only line-level), we adopt a mixed-granularity training strategy.
% The total training objective $\mathcal{L}$ is defined as the conditional sum of task-specific losses $\mathcal{L}_t$ for each granularity $t \in T = \{word, line, paragraph\}$:
%   \vspace{-0.3em}
% \begin{equation}
% \label{loss_full}
% \mathcal{L} = \sum_{t \in T} \omega_t \cdot \mathcal{L}_t 
% \vspace{-0.3em}
% \end{equation}
% Here, $\omega_t$ is a binary indicator set to 1 if the ground-truth for granularity $t$ is present in the current sample, and 0 otherwise. This ensures that each granularity-specific head is only supervised by its corresponding available ground-truth data.

% This training paradigm offers three key advantages. First, it allows Prism to leverage a wide range of public datasets, even those with limited or single-granularity annotations, for joint training. Second, our unified framework generates multi-granularity outputs in a single forward pass, significantly improving inference efficiency compared to training separate models. Finally, this approach enables our trained model to generate high-quality pseudo-labels for granularities absent from the original training data  (\textit{e.g.}, generating word-level detections on a line-level dataset), which we further demonstrate in the Appendix.

\noindent \textbf{Mixed-Granularity Training.} 
To effectively train our unified model on diverse datasets that often contain incomplete annotations (e.g., some datasets provide only word-level labels, while others provide only line-level labels), we adopt a mixed-granularity training strategy. The total training objective $L$ is defined as the conditional sum of task-specific losses $L_t$ for each granularity $t \in T = \{\text{word}, \text{line}, \text{paragraph}\}$:
\[
L = \sum_{t \in T} \omega_t \cdot L_t .
\]
Here, $\omega_t$ is a binary indicator set to 1 if the ground-truth for granularity $t$ is present in the current sample, and 0 otherwise. This conditional objective allows Prism to seamlessly leverage datasets with incomplete or single-granularity annotations, while producing multi-granularity predictions within a unified model. In addition, the trained model can infer missing granularities on datasets without complete hierarchical labels, making it useful for pseudo-label generation.

\noindent \textbf{Geometric Post-processing.} 
During post-processing, Prism reconstructs the hierarchy using box-level geometric inclusion. The parsing is performed in a bottom-up manner with two successive assignment steps: word-to-line and line-to-paragraph. For both steps, we use the same Box-IoS criterion. Specifically, given a lower-level text box $B_l$ and a higher-level text box $B_h$, we define their Box-IoS as
$$\mathrm{IoS}(B_l, B_h) = \frac{\mathrm{Area}(B_l \cap B_h)}{\mathrm{Area}(B_l)}$$
For each lower-level instance $B_l$, we first find the higher-level candidate with the maximum Box-IoS:
$$h^* = \arg\max_{h} \mathrm{IoS}(B_l, B_h), \qquad s^* = \max_{h} \mathrm{IoS}(B_l, B_h)$$
The assignment is then accepted only when the maximum score exceeds a threshold $\tau$:
$$\pi(B_l) = \begin{cases} B_{h^*}, & s^* \ge \tau, \\ \varnothing, & s^* < \tau \end{cases}$$

By estimating inclusion at the box level, Prism computes all pairwise IoS scores in each assignment step as an $N \times M$ matrix, where $N$ and $M$ denote the numbers of lower-level and higher-level instances, respectively. The row-wise maximum directly gives the parent assignments, reducing hierarchy parsing from words to lines and lines to paragraphs into two simple, parameter-free matrix operations.

\section{Experiments}

\subsection{Datasets}
 \textbf{HierText} \cite{long2022towards} is a large-scale dataset featuring hierarchical annotations for word, text-line, and paragraph positions. It contains 8,281 training, 1,724 validation, and 1,634 test images. 
\textbf{Total-Text} \cite{ch2017total} is a key arbitrary-shaped text benchmark, containing 1,000 training and 255 test images. \textbf{CTW1500} \cite{liu2019curved} is another arbitrary-shaped benchmark, providing 1,000 training and 500 test images annotated at the text-line level. \textbf{MSRA-TD500} \cite{yao2012detecting} is a multi-oriented dataset with 300 training and 200 test images. \textbf{ArT} \cite{chng2019icdar2019} 
is a large-scale  arbitrary-shaped text detection dataset, containing $5603$ training images and $4563$ test images.
\textbf{Synth150K} \cite{liu2020abcnet} is a synthetic pre-training dataset, comprising about 150K images. 
 \textbf{ICDAR 2013 (IC13)} \cite{karatzas2013icdar} is a horizontal dataset that contains $229$ training and $233$ test images.
 \textbf{ICDAR 2015 (IC15)} \cite{karatzas2015icdar} is a multi-oriented dataset with 1,000 training and 500 test images, labeled at the word-level.
\textbf{ICDAR17 MLT (MLT17)} \cite{nayef2017icdar2017} is a large multi-language dataset providing 9,000 images for training.
\textbf{TextOCR} \cite{singh2021textocr} is a large real-world scene text dataset, composed of 21,749 training, 3,153 validation, and 3,232 test images.

\subsection{Implementation Details}

We utilize a simplified ResNet50-DCN backbone, truncated after the C4 stage, and adopt the pixel-level regression head (Fig.~\ref{fig:head}(a)) as our default Expert Head. During training, we pre-train for 450K iterations on a mixture of Synth150K, IC13, IC15, MLT17, and TextOCR, using an ABCNet v2 recognition head for joint training. We use the AdamW optimizer (batch size 16) with an initial learning rate of 0.0001, which is decayed by a factor of 0.1 at 375K iterations. Then, we fine-tune for 300 epochs on each evaluated dataset.   
Data augmentations include \emph{RandomCrop}, \emph{RandomRotate}, \emph{RandomFlip}, and \emph{ColorJitter}. During inference, we resize the short sides of the input images to $704$, $1000$, $800$, $1600$, and $2048$ for CTW1500, Total-Text, MSRA-TD500, ArT, and HierText, respectively, while keeping the original aspect ratio. The IoS threshold $\tau$ is set to 0.3. Inference is performed on a single NVIDIA RTX 3090 GPU. For the HierText dataset, we follow \cite{long2022towards, ye2024hi} and adopt Panoptic Quality (PQ) and F-measure (F) as primary metrics to evaluate the detection and geometric layout performance.

\begin{table}[!tb]
\centering
\begin{minipage}{\linewidth}
\resulttableformat
\caption{\textbf{Scene text detection results on typical benchmarks.} All FPS values are uniformly measured using a single NVIDIA RTX 3090 GPU.}
\label{tab:det}
\begin{tabularx}{\linewidth}{l*{13}{Y}}
      \toprule \multirow{2}*{Method} & \multicolumn{3}{c}{MSRA-TD500} & \multicolumn{3}{c}{ICDAR19 ArT} & \multicolumn{3}{c}{Total-Text} &
      \multicolumn{4}{c}{CTW1500}  \\
    \cmidrule(l{0.2em}r{0.2em}){2-4} 
    \cmidrule(l{0.2em}r{0.2em}){5-7} 
    \cmidrule(l{0.2em}r{0.2em}){8-10} 
    \cmidrule(l{0.2em}r{0.2em}){11-14} 
       & R  & P & F & R  & P & F & R  & P & F  & R  & P & F &FPS \\
      \midrule

      PCR \cite{dai2021progressive} & 83.5 & 90.8 & 87.0  & 66.1 & 84.0 & 74.0 & 82.0 & 88.5 & 85.2 & 82.3 & 87.2 & 84.7 & \textbf{--} \\
      FSG \cite{tang2022few}       & $84.8$                 & $91.6$      & $88.1$  & \textbf{--} & \textbf{--} & \textbf{--} & $85.7$      & $90.7$       & $88.1$     & $82.4$  & $88.1$      & $85.2$   & \textbf{--}         \\  

     TextPMs \cite{zhang2022arbitrary}         & $87.0$                 & $91.0$      & $88.9$   & \textbf{--} & \textbf{--} & \textbf{--} & $87.7$      & $90.0$       & $88.8$     & $83.8$                 & $87.8$      & $85.7$    & $14.4$         \\ 
    DB++  \cite{liao2022real}        & $83.3$                 & $91.5$      & $87.2$  & \textbf{--} & \textbf{--} & \textbf{--}
   & $83.2$      & $88.9$       & $86.0$     & $82.8$                 & $87.9$      & $85.3$   & $\underline{38.3}$          \\
  TPSNet \cite{wang2022tpsnet}          & \textbf{--}                 & \textbf{--}      & \textbf{--}  
  & 73.3     & 84.3      & 78.4   
 & $86.8$      & $89.5$       & $88.1$        & $85.1$        & $87.7$      & $86.4$      & $17.9$         \\ 
  TextBPN++ \cite{zhang2023arbitrary}     & $86.8$                 & 93.7      & 90.1    
  & 77.1 & 84.5 & \underline{80.6} & $87.9$      & $92.4$   & 90.1      & $84.7$                 & $88.3$      & $86.5$    & $13.9$   \\
  DPText-DETR  \cite{ye2023dptext}    & \textbf{--}                 & \textbf{--}      & \textbf{--}   & 73.7 & 83.0  & 78.1  & $86.4$      & $91.8$       & $89.0$         & $86.2$     & \underline{91.7} 
 & $88.8$  & $14.8$  \\

 SRFormer \cite{bu2024srformer} & \textbf{--}  & \textbf{--}     & \textbf{--} & 73.5 & \textbf{86.1} & 79.3 & 87.9 & 91.5 & 89.7 & \textbf{89.8} & 89.4 & \underline{89.6} & 8.2 \\
 LRANet \cite{su2024lranet}   & $86.3$                 & $92.3$      & $89.2$   & \textbf{--}   & \textbf{--} & \textbf{--} & $87.8$      &$90.3$       & $89.0$      & $85.5$     & $89.4$        & $87.4$   & $37.2$ \\
  LayoutFormer  \cite{liang2024layoutformer}          & 88.3   & $92.0$       & 90.1    & \textbf{--}     & \textbf{--}      & \textbf{--}   & $85.0$      & $89.3$       & $87.1$  & $84.3$     & $88.2$ & $86.2$  & \textbf{--}  \\ 
  KAC \cite{zheng2024kernel} & 88.1 & 93.9 & 90.8  
  & \underline{77.3}  & 83.0 & 80.0  & 88.1 & \textbf{94.1} & 91.0  & 85.4 & 88.6 & 86.8 &\textbf{--} \\
FastTCM-TESTR \cite{yu2024turning} & \textbf{--} & \textbf{--} & \textbf{--} & \textbf{--} & \textbf{--} & \textbf{--} & \textbf{--} & \textbf{--} & 89.0 & \textbf{--} & \textbf{--} & 87.4 & \textbf{--} \\
DAT-DET \cite{wan2024towards} & 86.6 & \underline{95.1} & 90.7  & \textbf{--} & \textbf{--} & \textbf{--} & \underline{88.2} & \underline{94.0} & \underline{91.0} & \underline{89.3} & 89.3 & 89.3 & \textbf{--} \\
ERRNet \cite{su2025explicit} & 87.1 & 93.8 & 90.3 
& 75.5 & 84.1 & 79.6 & 87.3 & 92.6 & 89.9 & 87.9 & 91.0 & 89.4 & 31.5 \\
DocSAM \cite{li2025docsam} & \textbf{--} & \textbf{--} & \textbf{--}  & \textbf{--} & \textbf{--} & \textbf{--} & 82.6 & 72.1 & 77.0 & 88.1 & 80.5 & 84.2 & \textbf{--} \\
MTD \cite{han2025pull} & 88.0 & 94.6 & \underline{91.2} & \textbf{--} & \textbf{--} & \textbf{--} & 87.1 & 90.1 & 88.6 & 87.1 & 90.2 & 88.6 & \textbf{--} \\ 
OmniParser V2 \cite{yu2026omniparser}   & \textbf{--} & \textbf{--} & \textbf{--}  & \textbf{--} & \textbf{--} & \textbf{--} & 88.2 & 90.6 & 89.4 & 87.1 & 89.7  & 88.4    & \textbf{--} \\ 
 \midrule
 \rowcolor{resultrow}\textbf{Prism}   &   \textbf{89.5}  & \textbf{95.3} & \textbf{92.3}  & \textbf{77.9}   & \underline{84.6} & \textbf{81.1} & \textbf{89.6}      &92.9       & \textbf{91.2}    & 88.5  & \textbf{92.0}        & \textbf{90.2}   & \textbf{65.1} \\
      \bottomrule
    \end{tabularx}
\end{minipage}
\end{table}

\subsection{Comparison with State-of-the-art Methods}

\noindent \textbf{Single-granularity Text Detection.} We evaluate the performance of Prism on the most widely used benchmarks for single-granularity scene text detection, \textit{i.e.}, MSRA-TD500 and CTW1500 for text-line detection, and Total-Text and ArT for word detection. As shown in Table~\ref{tab:det}, Prism consistently demonstrates top-tier performance across these datasets. Specifically, on MSRA-TD500, our method outperforms the leading method MTD \cite{han2025pull} by 1.1\% (92.3\% vs. 91.2\%) in F-measure. On the arbitrary-shaped text benchmarks ArT and Total-Text, Prism also achieves SOTA performance with F-measures of 81.1\% and 91.2\%, respectively. Most importantly, Prism exhibits a significant advantage in inference speed. Benefiting from its minimalist design, it runs at 65.1 FPS on CTW1500, which is $1.7\times$ faster than LRANet and $7.9\times$ faster than SRFormer, establishing a new optimal trade-off between accuracy and efficiency.

\begin{table}[!htb]
\centering
\begin{minipage}{\linewidth}
\resulttableformat
\caption{\textbf{Hierarchical text detection results on the HierText dataset.}}
\label{table:hiertext_results}
\begin{tabularx}{\linewidth}{l*{12}{Y}} % 
        \toprule  
        & \multicolumn{6}{c}{Validation Set} & \multicolumn{6}{c}{Test Set} \\
        \cmidrule(lr){2-7} \cmidrule(lr){8-13}
        \multirow{1}{*}{Method} & \multicolumn{2}{c}{Word} & \multicolumn{2}{c}{Line} & \multicolumn{2}{c}{Para.} & \multicolumn{2}{c}{Word} & \multicolumn{2}{c}{Line} & \multicolumn{2}{c}{Para.} \\
        \cmidrule(lr){2-3} \cmidrule(lr){4-5} \cmidrule(lr){6-7} 
        \cmidrule(lr){8-9} \cmidrule(lr){10-11} \cmidrule(lr){12-13}
        & PQ & F & PQ & F & PQ & F & PQ & F & PQ & F & PQ & F \\
        \midrule 
        UniDec \cite{long2022towards} & 48.5 & 61.7 & 61.3 & 78.7 & 52.9 & 67.7 & 48.2 & 61.5 & 62.2 & 79.9 & 53.6 & 68.6 \\   
        LayoutFormer \cite{liang2024layoutformer} & \textbf{--} & \textbf{--} & \textbf{--} & \textbf{--} & \textbf{--} & \textbf{--} & 49.2 & 64.9 & 46.2 & 60.7 & 43.0 & 56.4 \\    
        Hi-SAM-B \cite{ye2024hi} & 59.2 & 77.5 & 62.9 & 81.4 & 55.7 & 72.6 & 59.7 & 78.3 & 63.3 & 82.2 & 54.5 & 71.2 \\
        Hi-SAM-L \cite{ye2024hi} & 62.6 & 81.0 & 65.6 & 84.0 & 58.8 & 75.8 & 63.1 & 81.8 & 66.2 & 84.9 & 57.6 & 74.5 \\    
        Omniparser V2 \cite{yu2026omniparser} & 60.0 & \textbf{--} & 63.4 & \textbf{--} & 55.9 & \textbf{--} & 61.6 & \textbf{--} & 64.5 & \textbf{--} & 55.2 & \textbf{--} \\  
        \midrule
        \rowcolor{resultrow}\textbf{Prism} & \textbf{75.3} & \textbf{90.1} & \textbf{70.6} & \textbf{85.9} & \textbf{67.6} & \textbf{82.2} & \textbf{75.8} & \textbf{90.4} & \textbf{71.1} & \textbf{86.0} & \textbf{66.8} & \textbf{81.4} \\
        \bottomrule
    \end{tabularx}
\end{minipage}
\end{table}

\noindent \textbf{Hierarchical Text Detection.} 
We compare Prism with existing frameworks on the challenging HierText test and validation sets (Table~\ref{table:hiertext_results}). Despite its minimalist design, Prism achieves a significant lead over existing methods that rely on complex architectures. On the test set, Prism surpasses the  SOTA OmniParser V2 by substantial margins across all hierarchical levels: +14.2\% in Word PQ (75.8 vs. 61.6), +6.6\% in Line PQ (71.1 vs. 64.5), and +11.6\% in Paragraph PQ (66.8 vs. 55.2). This gap stems primarily from three factors (with detailed ablations in Sec.~\ref{ablation}). First, hierarchical relationships can be naturally derived from detection outputs (words, lines, paragraphs) through simple post-processing. Unlike methods such as LayoutFormer \cite{liang2024layoutformer} that use relationship prediction modules and thus risk propagating cumulative errors, we obtain them by directly calculating the intersection-of-sets (IoS) between the minimum bounding boxes of the level-specific predictions, which is also highly efficient due to parallelization. Second, previous methods all adopt query-level heads; however, pixel-level heads have clear superiority over query-based ones for this task. Finally, our lightweight architecture allows us to efficiently adopt high-resolution inputs ($S=2048$). This is crucial because HierText contains a massive number of dense, small text instances that require high resolution for accurate detection. 
%Qualitative visualizations are provided in Fig.~\ref{fig:vis}.

\begin{table}[!t]
    \centering
\begin{minipage}{\linewidth}
\resulttableformat
\caption{\textbf{Pre-training}.
    Ablation study of pre-training setups on Total-Text and CTW1500. E2E training means adding a recognition head for joint training.}
\label{tab:ablation_pretrain}
\begin{tabular*}{\linewidth}{@{\extracolsep{\fill}}llcccccc@{}}
        \toprule
        \multirow{2}{*}{Pre-train Data} & \multirow{2}{*}{Method} & \multicolumn{3}{c}{Total-Text} & \multicolumn{3}{c}{CTW1500} \\
        \cmidrule(lr){3-5} 
        \cmidrule(lr){6-8} 
        & & R & P & F & R & P & F \\ 
        \midrule
        \multirow{2}{*}{None} 
            & LRANet & 85.7 & 90.5 & 88.1 & 84.9 & 89.1 & 86.9 \\
            & Ours   & 84.7 & 88.3 & \phantom{\qdecnew{.00}}86.5\qdecnew{1.6} & 83.5 & 86.1 & \phantom{\qdecnew{.00}}84.8\qdecnew{2.1} \\ 
        \midrule
        \multirow{2}{*}{+ Synth150K} 
            & LRANet & 87.8 & 90.3 & 89.0 & 85.5 & 89.4 & 87.4 \\
            & Ours   & 87.1 & 89.5 & \phantom{\qdecnew{.00}}88.3\qdecnew{0.7} & 85.6 & 87.5 & \phantom{\qdecnew{.00}}86.5\qdecnew{0.9} \\ 
        \midrule
        \multirow{2}{*}{\makecell[l]{\quad + IC13, IC15, MLT17}}
            & LRANet & 87.6 & 90.9 & 89.2 & 87.8 & 89.9 & 88.8 \\
            & Ours   & 88.1 & 90.0 & \phantom{\qdecnew{.00}}89.0\qdecnew{0.2} & 87.1 & 90.0 & \phantom{\qdecnew{.00}}88.5\qdecnew{0.3} \\ 
        \midrule
        \multirow{2}{*}{\quad \quad + TextOCR}
            & LRANet & 88.1 & 91.2 & 89.6 & 88.5 & 90.5 & 89.5 \\
            & Ours   & 88.3 & 90.9 & \phantom{\qeqnew{.00}}89.6\qeqnew{0.0} & 88.3 & 90.5 & \phantom{\qdecnew{.00}}89.4\qdecnew{0.1} \\ 
        \midrule
        \multirow{2}{*}{\quad \quad \quad + E2E Training}
            & LRANet & 89.7 & 92.3 & 91.0 & 88.8 & 91.7 & 90.2 \\ 
            & Ours   & 89.6 & 92.9 & \phantom{\qincnew{.00}}91.2\qincnew{0.2} & 88.5 & 92.0 & \phantom{\qincnew{.00}}90.2\qeqnew{0.0} \\ 
                
        \bottomrule
    \end{tabular*}
\end{minipage}
\par\addvspace{\floatsep}\nointerlineskip
\nextfloat
\begin{minipage}{\linewidth}
\resulttableformat
\caption{\textbf{Backbone Capacity.}
    Influence of different backbone networks. All models use the full pre-training setup (\textit{i.e.}, the E2E Training setting in Table~\ref{tab:ablation_pretrain}). * means truncation of C5. GFLOPs and FPS are evaluated at a short-side resolution of 704.}
\label{tab:ablation_backbone}
\begin{tabular*}{\linewidth}{@{\extracolsep{\fill}}llccccc@{}}
        \toprule
        \multirow{2}{*}{Backbone} & \multirow{2}{*}{Method} & \multirow{2}{*}{Params (M)} & \multirow{2}{*}{GFLOPs} & \multirow{2}{*}{FPS} & Total-Text & CTW1500 \\
        \cmidrule(lr){6-6} \cmidrule(lr){7-7}
        & & & & & F & F \\ 
        \midrule
        ResNet18 & LRANet  & 18.0 & 62.1 & \phantom{\qincnew{00.0}}65.4\phantom{\qincnew{00.0}} & 84.5 & 83.1 \\
        ResNet18* & Ours & 5.9  & 18.4 & \phantom{\qincnew{00.0}}158.6\qincnew{93.2} & \phantom{\qdecnew{0.0}}79.3\qdecnew{5.2} & \phantom{\qdecnew{0.0}}77.7\qdecnew{5.4} \\ 
        \midrule
        ResNet50  & LRANet & 31.0  & 81.9  & 40.5 & 87.8 & 86.5 \\
        ResNet50* & Ours & 11.8  & 34.1 & \phantom{\qincnew{00.0}}78.6\qincnew{38.1} & \phantom{\qdecnew{0.0}}85.2\qdecnew{2.6} & \phantom{\qdecnew{0.0}}83.7\qdecnew{2.8}  \\ 
        \midrule
        ResNet50-DCN  & LRANet & 32.7  & 97.2 & 37.2 & 91.2 & 90.3 \\
        ResNet50-DCN* & Ours & 12.8  & 48.3 & \phantom{\qincnew{00.0}}65.1\qincnew{27.9} & \phantom{\qdecnew{0.0}}91.2\qeqnew{0.0} & \phantom{\qdecnew{0.0}}90.2\qdecnew{0.1} \\ 
        \midrule
        ResNet101-DCN  & LRANet & 53.8 & 144.8  & 26.3 & 91.5 & 90.4 \\
        ResNet101-DCN* & Ours & 33.9  & 95.5  & \phantom{\qincnew{00.0}}37.1\qincnew{10.8} & \phantom{\qeqnew{0.0}}91.6\qincnew{0.1} & \phantom{\qeqnew{0.0}}90.4\qeqnew{0.0} \\ 
            
        \bottomrule
    \end{tabular*}
\end{minipage}
\end{table}

\subsection{Ablation Study}
\label{ablation}

\noindent \textbf{Pre-training.}
Compared to the leading real-time method LRANet \cite{su2024lranet}, Prism adopts a lighter head, discards the FPN, and truncates the backbone at C4. 
As shown in Table~\ref{tab:ablation_pretrain}, when training from scratch, Prism lags LRANet by $1.6\%$ in F-measure. This deficit arises because all representation pressure is concentrated onto the single C4 feature map, demanding sufficient pre-training to learn robust representations for varying text scales and shapes. Thus, with more pre-training data, the performance gap incrementally shrinks. Crucially, our model surpasses LRANet when utilizing End-to-End (E2E) Training (joint training with a recognition head), achieving an F-measure of $91.2\%$ compared to LRANet's $91.0\%$. This is attributed to the fine-grained semantic information introduced by E2E training, which compels the heavily tasked C4 feature to become more discriminative.  We further verify this finding on Vision Transformer (ViT) architectures using the challenging HierText dataset. As shown in Table~\ref{tab:rebuttal_vit_horizontal}(b), when using a ViT-B backbone without pre-training,
Prism noticeably underperforms LRANet. Yet, under the full pre-training setup, Prism closes this gap, matching or slightly exceeding LRANet across all hierarchical levels. Overall, these experiments demonstrate that rich pre-training (including both training data and training paradigms) is the essential prerequisite for our minimalist architecture.

\begin{table}[!t]
    \centering
\begin{minipage}{\linewidth}
\resulttableformat
\caption{\textbf{Backbone capacity and pre-training on ViT.}
(a) Backbone scaling with the full pre-training setup from Table~\ref{tab:ablation_pretrain}.
(b) Pre-training ablation using ViT-B.}
\label{tab:rebuttal_vit_horizontal}
% The change annotations use the extra right padding. Keep it outside the Para. header rule.
\newcommand*{\vitdeltapadding}{1.8em}
\newcommand*{\vitpararuletrim}{\dimexpr\cmidrulekern+\vitdeltapadding-\tabcolsep\relax}
\begin{minipage}[t]{.48\linewidth}
\centering
{\footnotesize\bfseries (a) Backbone  scaling\par}\smallskip
\begin{tabular*}{\linewidth}{@{\extracolsep{\fill}}llccc@{\hspace{\vitdeltapadding}}}
            \toprule
            \multirow{2}{*}{Backbone} & \multirow{2}{*}{Model} & Word & Line & Para. \\
            \cmidrule(lr){3-3}
            \cmidrule(lr){4-4}
            \cmidrule(lr{\vitpararuletrim}){5-5}
            & & PQ & PQ & PQ \\
            \midrule
            \multirow{2}{*}{ViT-T}
                & LRANet & 69.8 & 64.5 & 59.8 \\
                & Ours   & 66.7 & 61.1 & 55.6\rlap{\qdecnew{4.2}} \\
            \midrule
            \multirow{2}{*}{ViT-S}
                & LRANet & 73.4 & 67.7 & 63.5 \\
                & Ours   & 72.6 & 66.7 & 62.3\rlap{\qdecnew{1.2}} \\
            \midrule
            \multirow{2}{*}{ViT-B}
                & LRANet & 76.3 & 71.2 & 66.6 \\
                & Ours   & 76.3 & 71.4 & 66.7\rlap{\qincnew{0.1}} \\
            \bottomrule
        \end{tabular*}
\end{minipage}\hfill
\begin{minipage}[t]{.48\linewidth}
\centering
{\footnotesize\bfseries (b) Pre-training setups\par}\smallskip
\begin{tabular*}{\linewidth}{@{\extracolsep{\fill}}llccc@{\hspace{\vitdeltapadding}}}
            \toprule
            \multirow{2}{*}{Pretrain} & \multirow{2}{*}{Model} & Word & Line & Para. \\
            \cmidrule(lr){3-3}
            \cmidrule(lr){4-4}
            \cmidrule(lr{\vitpararuletrim}){5-5}
            & & PQ & PQ & PQ \\
            \midrule
            \multirow{2}{*}{None}
                & LRANet & 73.2 & 68.4 & 63.5 \\
                & Ours   & 71.8 & 66.9 & 62.0\rlap{\qdecnew{1.5}} \\
            \midrule
            \multirow{2}{*}{Synth150K}
                & LRANet & 74.9 & 69.8 & 65.1 \\
                & Ours   & 74.5 & 69.2 & 64.6\rlap{\qdecnew{0.5}} \\
            \midrule
            \multirow{2}{*}{Full}
                & LRANet & 76.3 & 71.2 & 66.6 \\
                & Ours   & 76.3 & 71.4 & 66.7\rlap{\qincnew{0.1}} \\
            \bottomrule
        \end{tabular*}
\end{minipage}
\end{minipage}
\par\addvspace{\floatsep}\nointerlineskip
\nextfloat
\begin{minipage}{\linewidth}
\resulttableformat
\caption{\textbf{Input size.} Performance comparison on HierText test set. ``\emph{S}:'' means the length of the shorter side, and the original aspect ratio is maintained.}
\label{table:size}
\begin{tabularx}{\linewidth}{l*{10}{Y}}
        \toprule  
        \multirow{2}{*}{Scale} & \multicolumn{2}{c}{Word} & \multicolumn{2}{c}{Line} & \multicolumn{2}{c}{Paragraph} & \multicolumn{3}{c}{Latency (ms)} & \multirow{2}{*}{FPS} \\
        \cmidrule(lr){2-3} \cmidrule(lr){4-5} \cmidrule(lr){6-7} \cmidrule(lr){8-10}
        & PQ & F & PQ & F & PQ & F & Backb. & Head & Post. & \\
        \midrule 
        \emph{S}: 1024 & 65.2 & 82.5 & 63.2 & 79.6 & 60.9 & 76.8 & \textbf{26.7} & \textbf{4.8} & \textbf{2.1} & \textbf{29.7} \\ 
        \emph{S}: 1536 & 72.6 & 88.5 & 69.6 & 84.8 & 66.6 & 81.0 & 61.3 & 9.8 & 2.2 & 13.6 \\  
        \emph{S}: 2048 & \textbf{75.8} & \textbf{90.4} & \textbf{71.1} & \textbf{86.0} & \textbf{66.8} & \textbf{81.4} & 104.7 & 16.1 & 2.3 & 8.1 \\ 
        \bottomrule
    \end{tabularx}
\end{minipage}
\par\addvspace{\floatsep}\nointerlineskip
\nextfloat
\begin{minipage}{\linewidth}
\resulttableformat
\caption{\textbf{Head design.} Comparisons of different lightweight heads in Fig.~\ref{fig:head} on HierText test set.}
\label{table:head}
\begin{tabularx}{\linewidth}{l*{6}{Y}c}
        \toprule
        \multirow{2}{*}{Type} & \multicolumn{2}{c}{Word} & \multicolumn{2}{c}{Line} & \multicolumn{2}{c}{Para.} & \multicolumn{1}{c}{Latency (ms)} \\
        \cmidrule(lr){2-3} 
        \cmidrule(lr){4-5} 
        \cmidrule(lr){6-7} 
        \cmidrule(lr){8-8}
        & PQ & F & PQ & F & PQ & F & Head \\
        \midrule
        
        Query-Seg. & 65.7 & 82.8 & 60.5 & 77.3 & 55.4 & 71.2 & 27.2 \\
        Query-Reg. & 67.9 & 85.1 & 63.3 & 80.5  & 58.2 & 74.9 & 24.8 \\
        Pixel-Seg. & 73.1 & 88.5 & 67.3 & 83.2 & 62.5 & 77.7 & 20.6 \\ 
       \rowcolor{resultrow} Pixel-Reg. & \textbf{75.8}    & \textbf{90.4}  & \textbf{71.1} & \textbf{86.0}  & \textbf{66.8} & \textbf{81.4}  & \textbf{16.1} \\
        \bottomrule
     \end{tabularx}
\end{minipage}
\end{table}

\noindent \textbf{Backbone Capacity.}
We evaluate our hypothesis that a powerful backbone is the core prerequisite for minimalist design across both CNN and Transformer architectures. As shown in Table~\ref{tab:ablation_backbone}, when employing lighter backbones such as ResNet18 and ResNet50, Prism's F-measure significantly lags LRANet. This deficit is mainly due to insufficient feature representation ability in these backbones, where LRANet's multi-scale neck and multi-layer heads act as compensatory mechanisms to cover this shortfall. However, as the backbone is enhanced, these compensatory mechanisms become redundant, as our method achieves comparable performance with LRANet when using ResNet50-DCN. Crucially, our minimalist architecture maintains a substantial efficiency advantage, requiring less than half the parameters and GFLOPs of LRANet while delivering a 1.7$\times$ inference speedup (65.1 vs. 37.2 FPS). Interestingly, further scaling to ResNet101-DCN yields negligible gains for both methods, suggesting ResNet50-DCN already provides sufficient feature discriminativeness for the purely perceptual task of text detection. This scaling law generalizes to Vision Transformers (Table~\ref{tab:rebuttal_vit_horizontal}(a)): while Prism noticeably trails LRANet with a ViT-T backbone ($66.7\%$ vs. $69.8\%$ Word PQ), scaling to ViT-B entirely eliminates this gap, confirming that a capable backbone intrinsically renders sophisticated necks obsolete.

\noindent \textbf{Expert Head.}
In Table~\ref{table:head}, we evaluate four 
lightweight head archetypes discussed in Sec. \ref{method:head}. To accommodate the massive number of dense text instances in HierText, we set the number of queries in query-level heads to 500, 300, and 100 for word, line, and paragraph detection, respectively. However, query-level heads consistently and substantially underperform their pixel-level counterparts. This is primarily because optimizing such a large set of queries for tiny, densely packed instances remains a significant challenge. Moreover, the global interaction of query-level heads offers limited benefits for this purely perceptual task, as text detection mainly relies on local patterns rather than contextual reasoning. Such interaction may be more suitable for logical layout analysis (\textit{e.g.}, distinguishing footers from body text). In contrast, pixel-level heads can fully leverage the explicit local spatial priors on the feature map, which is better suited to text detection. Of the two pixel-level variants, the regression head has a clear advantage, achieving the highest performance across all granularities: Word (75.8 PQ), Line (71.1 PQ), and Paragraph (66.8 PQ). This superiority is mainly because of its holistic instance perception. Compared to the complex bottom-up pixel grouping required by Pixel-Seg, holistic regression is a more natural and effective paradigm for predicting boundaries directly from the highly abstract C4 feature map. Consequently, we adopt Pixel-Reg as the default head instantiation for Prism.

% \noindent \textbf{Hierarchy Parsing.} 
% As Table~\ref{table:hierarchy} shows, due to hierarchical inclusion being naturally encoded in spatial layouts, simple geometric post-processing yields inherently higher accuracy than a strong learnable module (detailed in the Appendix), which is redundant and prone to cumulative errors. However, since text instances are arbitrarily shaped, directly computing exact polygon intersections (Polygon-level IoS) is highly time-consuming. By adopting the Box-level IoS approximation, the accuracy loss is practically negligible (\textit{e.g.}, only a 0.1\% drop in Line PQ). More importantly, this approximation transforms complex geometric calculations into  parallelized matrix operations, drastically boosting the processing speed (reducing latency to merely 1.2 ms). 
%This demonstrates that our Box-level geometric inclusion is the optimal and most efficient approach for hierarchical structural parsing.

\begin{table}[!t]
    \centering
% Half the distance between the two header baselines, including the short rule.
\newcommand*{\measureheadergap}{%
  \edef\halfheadergap{\the\dimexpr
    (\arraystretch\baselineskip+\aboverulesep+\cmidrulewidth+\belowrulesep)/2\relax}}
\begin{minipage}{\linewidth}
\resulttableformat
\renewcommand{\arraystretch}{1.28}
\measureheadergap
\caption{\textbf{Hierarchy parsing.} Comparisons of different hierarchical relationship parsing strategies on HierText test set.}
\label{table:hierarchy}
\begin{tabularx}{\linewidth}{l*{6}{Y}c}
        \toprule
        \raisebox{-\halfheadergap}[0pt][0pt]{Hierarchy Parsing} & \multicolumn{2}{c}{Word} & \multicolumn{2}{c}{Line} & \multicolumn{2}{c}{Para.} & \multicolumn{1}{c}{Latency (ms)} \\
        \cmidrule(lr){2-3} 
        \cmidrule(lr){4-5} 
        \cmidrule(lr){6-7} 
        \cmidrule(lr){8-8}
        & PQ & F & PQ & F & PQ & F & Parsing Time \\
        \midrule
        Learnable Module & 74.7 & 89.2 & 69.1 & 83.9 & 64.7 & 79.2 & 8.5 \\
        Polygon-level IoS & \textbf{75.8} & \textbf{90.4} & \textbf{71.3} & \textbf{86.2} & \textbf{66.9} & \textbf{81.6} & 32.4 \\
     \rowcolor{resultrow}   Box-level IoS & \textbf{75.8} & \textbf{90.4} & 71.1 & 86.0 & 66.8 & 81.4 & \textbf{1.2} \\
        \bottomrule
     \end{tabularx}
\end{minipage}
\par\addvspace{\dimexpr\floatsep+3pt\relax}\nointerlineskip
\nextfloat
\begin{minipage}[t]{.57\linewidth}
\vspace{0pt}
\caption{\textbf{Unified vs. Separate Paradigms.} Comparison on the HierText test set. ``Separate'' denotes adopting three independent models for each granularity detection.}
\label{table:united}
\end{minipage}\hfill
\begin{minipage}[t]{.40\linewidth}
\vspace{0pt}
\nextfloat
\caption{Comparison of polygon-level and box-level IoU computation for NMS on a highly curved subset of CTW1500.}
\label{tab:curved_box_ablation}
\end{minipage}\par\nointerlineskip
\begin{minipage}[t]{.57\linewidth}
\vspace{0pt}
\resulttableformat
\measureheadergap
\begin{tabularx}{\linewidth}{l*{6}{Y}}
        \toprule    
        \raisebox{-\halfheadergap}[0pt][0pt]{Method} & \multicolumn{2}{c}{Word} & \multicolumn{2}{c}{Line} & \multicolumn{2}{c}{Para.}  \\
        \cmidrule(lr){2-3} 
        \cmidrule(lr){4-5} 
        \cmidrule(lr){6-7} 
        & PQ & F & PQ & F & PQ & F  \\
        \midrule 
        Separate & 75.6 & 90.2 & 70.8 & 84.2 & 66.5 & 79.8 \\         
       \rowcolor{resultrow} Unified  & \textbf{75.8} & \textbf{90.4} & \textbf{71.1} & \textbf{86.0} & \textbf{66.8} & \textbf{81.4} \\
        \bottomrule
     \end{tabularx}
\end{minipage}\hfill
\begin{minipage}[t]{.40\linewidth}
\vspace{0pt}
\resulttableformat
\begin{tabularx}{\linewidth}{l*{3}{Y}}
\toprule
IoU & R & P & F  \\
\midrule
Polygon-level & 85.3 & 87.8 & 86.5  \\
\rowcolor{resultrow} Box-level     & 85.1 & 87.8 & 86.4 \\
\bottomrule
\end{tabularx}
\end{minipage}
\end{table}

\begin{figure}[!t]
    \centering
    \begin{minipage}{.66\linewidth}
        \centering
        \useembeddedpdfgroup
        \includegraphics[width=\linewidth]{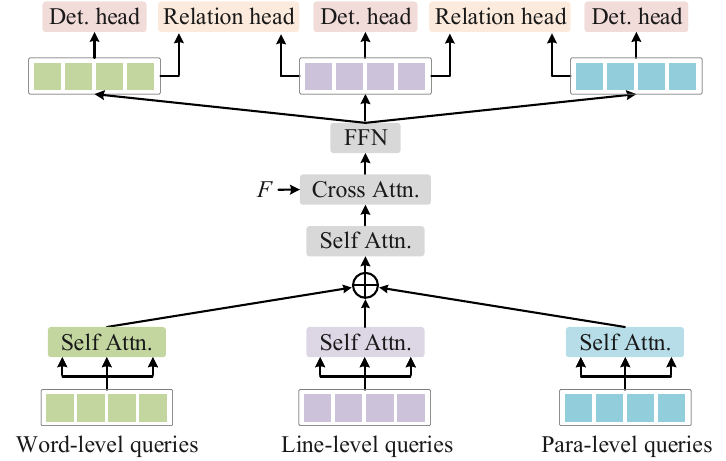}
        \caption{Detailed architecture of the learnable relation module used for comparison against our parameter-free geometric hierarchy parsing.}
        \label{fig:relation_module}
    \end{minipage}
\end{figure}

\begingroup
\clubpenalty=10000\relax
\noindent \textbf{Hierarchy Parsing.} 
As Table~\ref{table:hierarchy} shows, due to hierarchical inclusion being naturally encoded in spatial layouts, simple geometric post-processing yields inherently higher accuracy than a strong learnable module (illustrated in Fig.~\ref{fig:relation_module}). Specifically, we construct this strong baseline to explicitly model the structural hierarchy. It first selects top-scoring initial predictions and constructs queries by injecting level-specific embeddings and positional encodings. These queries are then iteratively refined through a complex decoding block repeated 6 times. Each block comprises independent self-attention within each granularity, unified self-attention to model cross-granularity context, and cross-attention interacting with the backbone feature map. Finally, the features of adjacent levels (\textit{e.g.}, word-line and line-paragraph pairs) are fused via element-wise multiplication and processed by an MLP-based relation head to explicitly predict their affiliations. However, such an intricate design is redundant and prone to cumulative errors.\par
\endgroup

\enlargethispage{.7\baselineskip}
Instead of relying on heavyweight learnable modules, we leverage geometric calculations. However, since text instances are arbitrarily shaped, directly computing exact polygon intersections (Polygon-level IoS) is highly time-consuming. By adopting the Box-level IoS approximation, the accuracy loss is practically negligible (\textit{e.g.}, only a 0.1\% drop in Line PQ). To further validate the effectiveness of the box-level approximation, we compare polygon-level and box-level IoU computation in NMS on a highly curved subset of CTW1500, following the subset definition in \cite{su2025explicit}. As shown in Table~\ref{tab:curved_box_ablation}, box-level IoU achieves nearly identical detection performance to polygon-level IoU, indicating that the box approximation introduces negligible accuracy loss even in this challenging setting. More importantly, this approximation transforms complex geometric calculations into parallelized matrix operations, substantially improving the processing speed and reducing latency to 1.2 ms.

\noindent \textbf{Unified vs. Separate.}
In Table~\ref{table:united}, we compare our unified learning paradigm against training three independent models for word, line, and paragraph detection, respectively. The unified model consistently outperforms the separate baseline across all metrics. We attribute this to two factors: 1) Hierarchical Synergy: Text entities exhibit inherent physical inclusion constraints (\textit{e.g.}, lines encompass words). Joint training allows the model to implicitly internalize these constraints, utilizing high-level context to verify and refine low-level detections. 2) Geometric Consistency: Since layout analysis depends heavily on the spatial alignment between granularities, sharing a unified feature encoder ensures that predictions for words and lines are spatially coherent, thereby reducing matching errors. Furthermore, the unified framework is inherently more concise and parameter-efficient.

\begin{table}[!t]
\centering
\begin{minipage}{.90\linewidth}
\resulttableformat
\caption{\textbf{Backbone feature selection.} Comparisons of different single-scale backbone feature maps on the HierText test set. $C_3$, $C_4$, and $C_5$ represent features at 1/8, 1/16, and 1/32 scales, respectively.}
\label{table:feature_map}
\begin{tabularx}{\linewidth}{l*{7}{Y}}
        \toprule
        \multirow{2}{*}{Feat.} & \multicolumn{2}{c}{Word} & \multicolumn{2}{c}{Line} & \multicolumn{2}{c}{Para.} & \multirow{2}{*}{FPS} \\
        \cmidrule(lr){2-3} 
        \cmidrule(lr){4-5} 
        \cmidrule(lr){6-7} 
        & PQ & F & PQ & F & PQ & F &  \\
        \midrule
        $C_3$  & 75.0 & 89.6 & 69.6 & 83.7 & 62.2 & 74.9 & 8.0 \\
     \rowcolor{resultrow}   $C_4$ & \textbf{75.8} & \textbf{90.4} & \textbf{71.1} & \textbf{86.0} & \textbf{66.8} & \textbf{81.4} & \textbf{8.1} \\
        $C_5$ & 68.0 & 86.1 & 64.8 & 82.0 & 61.1 & 77.3 & 7.4 \\
        \bottomrule
     \end{tabularx}
\end{minipage}
\end{table}

\noindent \textbf{Choice of Backbone Feature Map.}
In Table~\ref{table:feature_map}, we investigate the optimal single-scale feature map from the ResNet50-DCN backbone by comparing the $C_3$, $C_4$, and $C_5$ outputs. As observed, the high-resolution $C_3$ feature map yields a lower paragraph-level F-measure due to its limited receptive field. Conversely, the $C_5$ feature map suffers from a lower word-level F-measure owing to its low spatial resolution. The $C_4$ feature map effectively balances resolution and receptive field, achieving the highest F-measure across all granularities. Interestingly, while $C_5$ achieves a higher paragraph F-measure than $C_3$ (77.3 vs. 74.9), its paragraph Panoptic Quality (PQ) is actually lower (61.1 vs. 62.2). This discrepancy occurs because $C_5$'s low resolution drastically degrades word-level detection, which subsequently propagates errors up the bottom-up hierarchical parsing.
Additionally, extracting $C_5$ requires deeper backbone computations, whereas $C_3$ imposes a heavy computational burden on the detection heads due to its large spatial dimensions. As a result, the centrally positioned $C_4$ achieves the fastest inference speed. In summary, $C_4$ serves as the optimal compromise between accuracy and efficiency.

\begin{figure}[!t]
    \centering
    \begin{minipage}{\linewidth}
        \centering
        {\includegraphics[width=\linewidth]{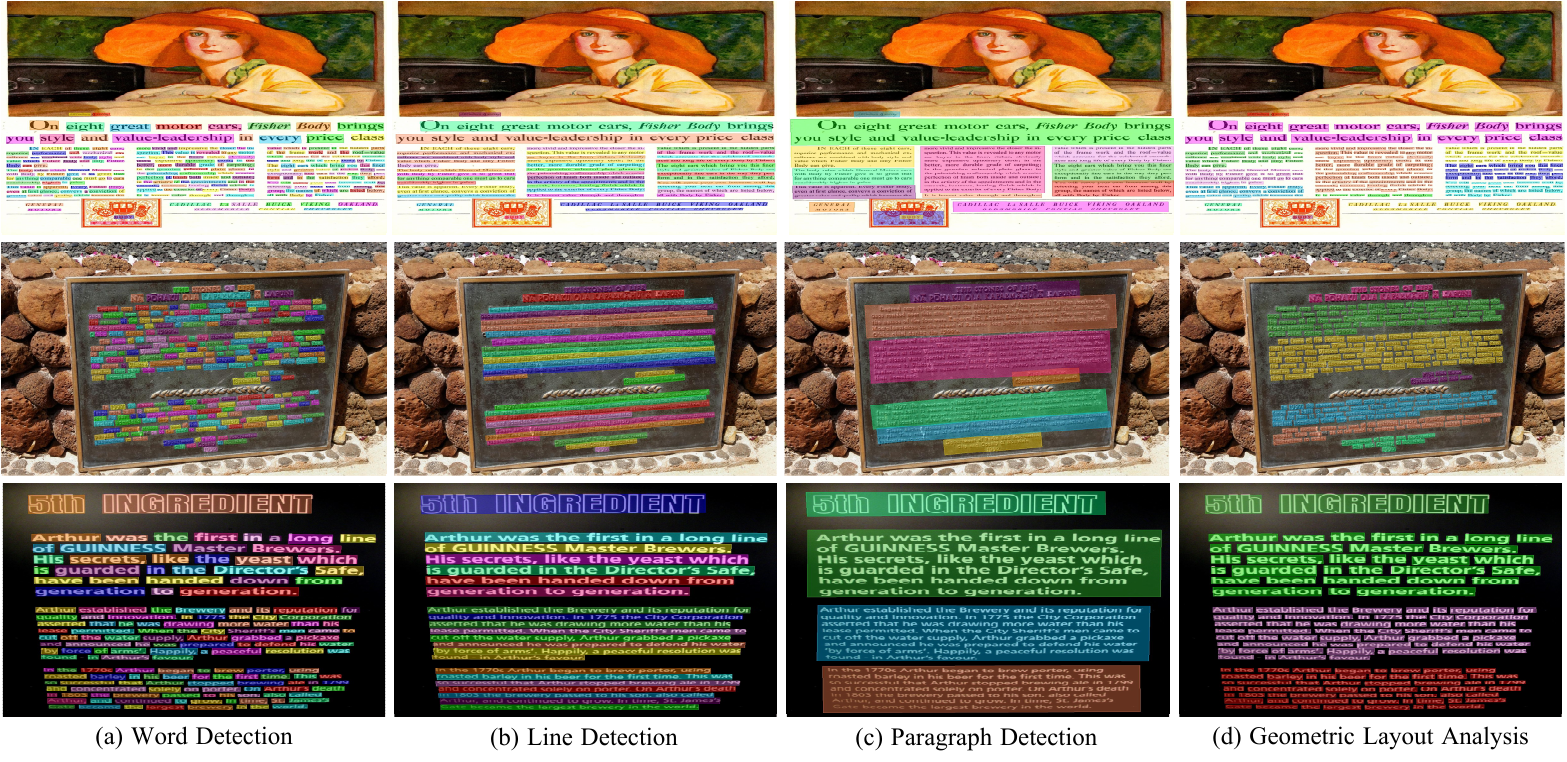}}
        \caption{\textbf{Visualization results of Prism on HierText test set.}}
        \label{fig:vis}
    \end{minipage}
\par\addvspace{\floatsep}\nointerlineskip
\nextfloat
    \begin{minipage}{\linewidth}
        \centering
        \includegraphics[width=\linewidth]{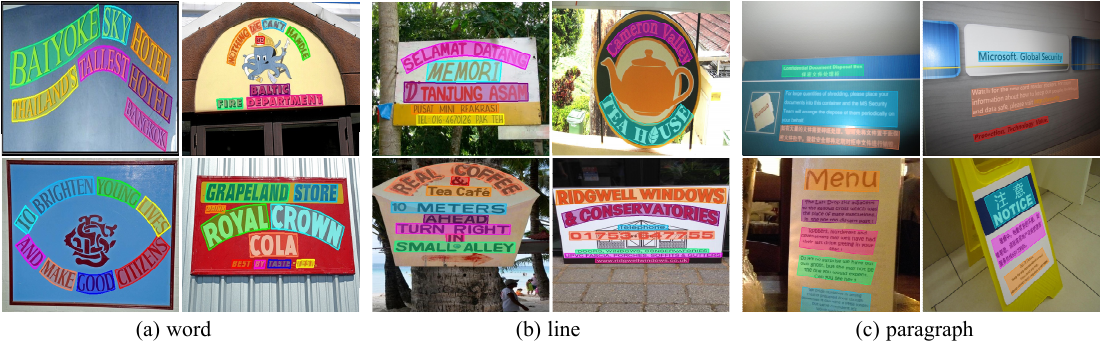}
        \caption{\textbf{Visualization of multi-granularity pseudo-labels generated by Prism.} Note that these datasets do not have annotations for these specific granularities.}
        \label{fig:vis3}
    \end{minipage}
\end{figure}

\begingroup
\looseness=-1\relax
\noindent \textbf{Input Image Size.}
In Table~\ref{table:size}, we investigate the impact of input image scale. As the short-side length increases from 1,024 to 2,048, performance improves substantially, especially for the Word PQ metric (rising from 65.2\% to 75.8\%). This is mainly because the HierText dataset contains a vast number of small-scale text instances. Increasing feature resolution is indispensable for accurately extracting such tiny text features. In this regard, our minimalist architecture holds a significant computational advantage compared to other methods. \par
\endgroup

\subsection{Visualization Analysis}

To intuitively demonstrate the effectiveness, robustness, and generalization capability of Prism, we conduct a comprehensive qualitative analysis across multiple benchmarks.

\begin{figure}[!t]
    \centering
    \begin{minipage}{\linewidth}
        \centering
        \includegraphics[width=\linewidth]{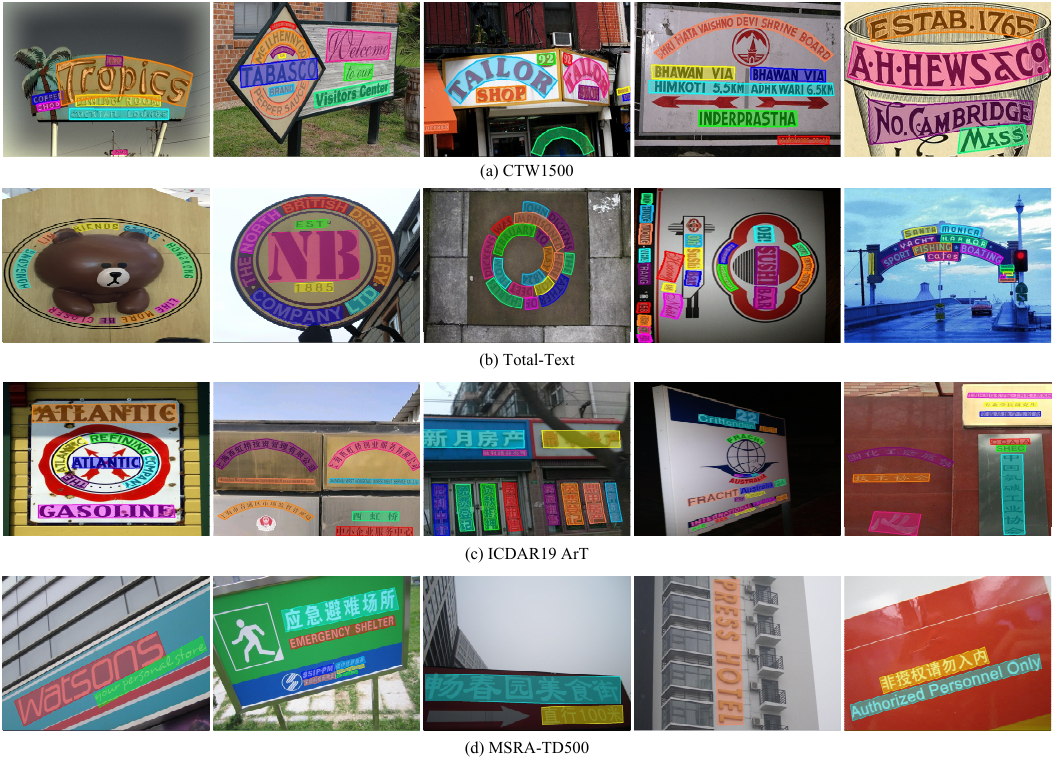}
        \caption{\textbf{Visualization results of Prism on scene text detection benchmarks:} CTW1500, Total-Text, ICDAR19 ArT, and MSRA-TD500.}
        \label{fig:vis2}
    \end{minipage}
\end{figure}

As illustrated in Fig.~\ref{fig:vis}, we first visualize the multi-granularity detection results on the HierText test set. In complex document layouts or natural scenes with dense, small text instances, Prism accurately localizes visual text across words, lines, and paragraphs, and parses their topological inclusion relationships via the proposed Box-level IoS algorithm. This demonstrates that minimalist geometric post-processing can effectively replace heavy relation-reasoning modules.

\begingroup
\widowpenalty=10000\relax
Furthermore, we evaluate Prism on standard single-granularity scene text detection benchmarks, including CTW1500, Total-Text, ICDAR19 ArT, and MSRA-TD500. As shown in Fig.~\ref{fig:vis2}, our method accurately detects text instances that exhibit challenging attributes, such as extreme length, high curvature, vertical orientation, or tiny scales. In particular, on CTW1500, tiny text instances are precisely localized even at a short-side resolution of 704. This robust performance demonstrates that our single-scale feature representation captures sufficient discriminative details to handle diverse shapes and scales without requiring multi-scale~feature~pyramids.\par
\endgroup

Benefiting from our mixed-granularity training strategy, Prism exhibits strong generalization capabilities. While many existing datasets only provide single-granularity annotations, Prism can seamlessly predict the missing hierarchical levels. As illustrated in Fig.~\ref{fig:vis3}, it can generate precise word-level boundaries for highly stylized and dense text, accurately group scattered elements into coherent lines, and successfully cluster multiple lines into logical paragraphs despite complex background interference.  This capability significantly enhances the practical value of our model, enabling the automatic refinement of incomplete annotations in real-world applications.

\FloatBarrier
\begin{samepage}
\section{Conclusion}
In this paper, we propose Prism, a minimalist hierarchical text detection framework. By systematically re-evaluating the necessity of complex architectural components, we demonstrate that a single-scale feature map, empowered by a sufficiently powerful and extensively pre-trained backbone, can provide a competitive and efficient alternative to traditional multi-scale necks and heavy heads. Moreover, our empirical study of four lightweight head archetypes concludes that a simple pixel-level regression head provides the best accuracy-efficiency trade-off for this setting. To parse the multi-granularity relationships, Prism reconstructs the structural hierarchy through a simple yet highly efficient geometric inclusion algorithm. Given its extreme simplicity, we hope that Prism can naturally serve as a plug-and-play detection extension for emerging large vision-language models (VLMs) and encourages future research to shift focus from intricate architectural engineering toward a deeper exploration of backbone networks and pre-training.

\end{samepage}

% The template's plainnat format, with proceedings issue numbers preserved.
\bibliographystyle{plainnat-prism}
\bibliography{IEEEabrv,reference}

\end{document}